\documentclass[letterpaper, 10 pt, conference]{ieeeconf}

\IEEEoverridecommandlockouts                              %

\usepackage[utf8]{inputenc}
\usepackage{hyperref}
\usepackage[pdftex,usenames,dvipsnames]{xcolor}  %

\usepackage{graphicx}

\usepackage{amsmath}
\usepackage{amsfonts}

\usepackage{subcaption}
\usepackage{multirow}

\allowdisplaybreaks
\usepackage{siunitx}
\graphicspath{{figures/},{data/Figures/},{}}

\newif\ifmargincomments %
\margincommentsfalse

\ifmargincomments
\newcommand{\federico}[2][1=]{\todo[inline,linecolor=magenta,backgroundcolor=magenta!25,bordercolor=magenta,#1]{#2}}
\newcommand{\tiago}[2][1=]{\todo[inline,linecolor=yellow,backgroundcolor=yellow!25,bordercolor=yellow,#1]{#2}}
\newcommand{\pending}[2][1=]{\todo[inline,linecolor=red,backgroundcolor=red!25,bordercolor=red,#1]{#2}}

\definecolor{mypink}{rgb}{0.858, 0.188, 0.478}
\newcommand{\mss}[1]{{\color{mypink}{#1}}}
\newcommand{\rev}[1]{{\color{blue}#1}}
\else
\newcommand{\federico}[2][1=]{}
\newcommand{\tiago}[2][1=]{}
\newcommand{\pending}[2][1=]{}

\newcommand{\mss}[2][1=]{}
\newcommand{\rev}[1]{#1}
\fi

\newif\ifjournalv
\journalvtrue
\ifjournalv
\usepackage{flushend}
\fi

\newcommand{\tasks}{\ensuremath{\mathbb{T}}}
\newcommand{\agents}{\ensuremath{\mathbb{N}}}
\newcommand{\requiredtasks}{\ensuremath{\mathbb{R}}}
\newcommand{\optionaltasks}{\ensuremath{\mathbb{O}}}
\newcommand{\reward}[1]{\ensuremath{r(#1)}}

\newcommand{\timescale}{\ensuremath{T}}
\newcommand{\task}{\ensuremath{t}}
\newcommand{\agent}{\ensuremath{i}}
\newcommand{\taskcomputationload}[2]{\ensuremath{c_{#1#2}}}
\newcommand{\taskpower}[2]{\ensuremath{p_{#1#2}}}
\newcommand{\taskproductsize}[1]{\ensuremath{d_{#1}}}
\newcommand{\taskchildren}[1]{\ensuremath{R({#1})}}
\newcommand{\linkcomputationloadout}[2]{\ensuremath{c^{\text{out}}_{#1#2}}}
\newcommand{\linkcomputationloadin}[2]{\ensuremath{c^{\text{in}}_{#1#2}}}
\newcommand{\linkenergyout}[2]{\ensuremath{e^{\text{out}}_{#1#2}}}
\newcommand{\linkenergyin}[2]{\ensuremath{e^{\text{in}}_{#1#2}}}
\newcommand{\maxthroughput}[2]{\ensuremath{\overline{b}_{#1,#2}}}
\newcommand{\links}{\ensuremath{\mathbb{E}}}
\newcommand{\lslatency}[2]{\ensuremath{l_{#1,#2}}}
\newcommand{\maxcomputationload}[1]{\ensuremath{\overline{c}_{#1}}}
\newcommand{\maxlatency}[2]{\ensuremath{\overline{l}_{#1,#2}}}

\newif\ifrelaxedv  %
\relaxedvfalse
\ifrelaxedv
\newcommand{\jvspace}[1]{}
 \else
\newcommand{\jvspace}[1]{\vspace{#1}}
\fi

\title{The Pluggable Distributed Resource Allocator (PDRA): \\a Middleware for Distributed Computing in Mobile Robotic Networks}
\author{Federico Rossi, Tiago Stegun Vaquero, Marc Sanchez Net et al }

\author{Federico Rossi$^{*}$, Tiago Stegun Vaquero$^{*}$, Marc Sanchez-Net, Maíra Saboia da Silva, and Joshua Vander Hook%
\thanks{$^{*}$ denotes equal contribution.}
\thanks{Authors affiliation: Jet Propulsion Laboratory, California Institute of Technology, Pasadena, CA, 91109, {\tt \{federico.rossi, tiago.stegun.vaquero, marc.sanchez.net, maira.saboia.da.silva, hook\}@jpl.nasa.gov}.}
}

\begin{document}

\maketitle

\begin{abstract}
We present the Pluggable Distributed Resource Allocator (PDRA), a middleware for distributed computing in heterogeneous mobile robotic networks.
PDRA enables autonomous robotic agents to share computational resources for computationally expensive tasks such as localization and path planning.
It sits between an existing single-agent planner/executor and existing computational resources (e.g. ROS packages), intercepts the executor’s requests and, if needed, transparently routes them to other robots for execution.
PDRA is \emph{pluggable}: it can be integrated in an existing single-robot autonomy stack with minimal modifications.
Task allocation decisions are performed by a mixed-integer programming algorithm, solved in a shared-world fashion, that models CPU resources, latency requirements, and multi-hop, periodic, bandwidth-limited network communications; the algorithm can minimize overall energy usage or maximize the reward for completing optional tasks.
Simulation results show that PDRA can reduce energy and CPU usage by over 50\% in representative multi-robot scenarios compared to a naive scheduler; runs on embedded platforms; and performs well in delay- and disruption-tolerant networks (DTNs). PDRA is available to the community under an open-source license.
\end{abstract}

\section{Introduction}

Multi-robot systems have been increasingly investigated in domains where single-robot missions have dominated for decades, including space, underwater and subterranean exploration.
Heterogeneous architectures are especially promising: for instance, heterogeneous robots can explore rough terrains or caves with a multitude of mobility modalities (e.g., drones, rovers, climbers), and orbiter-lander architectures can enable agile science and share communication and computation resources in planetary exploration. 

Crucially, mobile multi-robot systems can share \emph{computational} tasks, especially if they have heterogeneous computation capabilities -- for instance, a ground rover can offload expensive computations to an orbiting asset, or to a nearby ``base station''. %
Previous work \cite{hook-i-SAIRAS-2018,hook-et-al-icaps19} has shown that computation sharing in robotic systems with heterogeneous computing capabilities can lead to significant increases in system-level performance and science returns.

Realizing this computation-sharing paradigm within a multi-robot system requires not only algorithmic tools to allocate tasks (e.g., automated planning and scheduling tools), but also \emph{software architectures} that enable transparent sharing of computational workflows through available communication channels. Furthermore, it is especially desirable for such architectures to be seamlessly integrated with existing mission executives and existing robotics software libraries (e.g., the Robot Operating System, or ROS, ecosystem), thus enabling \emph{existing} autonomy architectures to adopt computation sharing with minimal changes.

\begin{figure}
	\centering
	\includegraphics[width=1.0\linewidth]{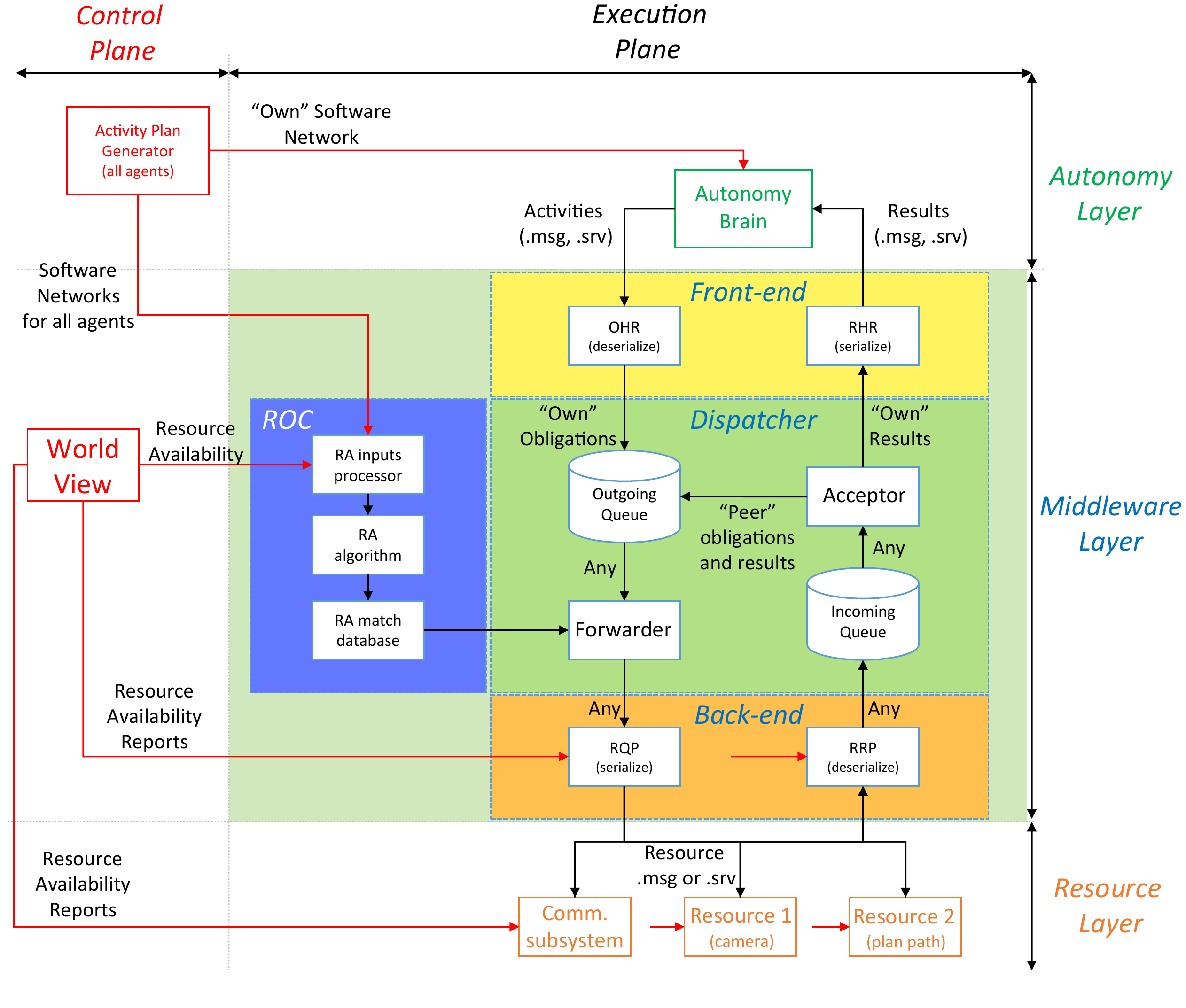}
	\caption{PDRA high-level architecture and its supporting processes.}
	\label{fig:pdra_simplified_architecture}
\end{figure}

To realize this paradigm, in this paper we propose a \emph{plug-in middleware} to transparently distribute computational tasks across a network of robotic agents with time-varying network connectivity. The middleware, shown in Figure \ref{fig:pdra_simplified_architecture}, is called the \textit{Pluggable Distributed Resource Allocator} (PDRA) and it is designed to:
\begin{itemize}
\item easily integrate with existing planning and execution systems and existing computation libraries;
\item be compatible with ROS (in particular with the ROS Action library) and also be easily portable to other robotics frameworks (e.g., the F-Prime flight software framework~\cite{f-prime-website});
\item operate in both time-invariant networks and Delay/Disruption Tolerant Networks (DTNs) \cite{burleigh2003dtn};
\item be open-source and available to the robotics community;
\end{itemize}

To allocate tasks to robotic agents, we propose a novel mixed-integer linear programming task allocation algorithm designed for networks with \emph{periodic} connectivity (e.g., orbiter-rover and data muling scenarios). We show that the algorithm can solve task allocation problems with tens of robots and hundreds of tasks in seconds, even on embedded computing platforms. For networks with non-periodic network connectivity, the PDRA architecture can readily accommodate other state-of-the-art scheduling and task allocation algorithms.

Our implementation of PDRA is available under the permissive Apache open-source license at \url{https://github.com/nasa/mosaic}.

\subsection{Literature Review}

\paragraph{Middlewares}
A number of middlewares for robotic systems support distributed or concurrent processing, communication, and message passing between agents. Examples of these include the popular \rev{ROS \cite{quigley2009ros}, its successor ROS2, ZeroMQ \cite{hintjens2013zeromq}, F-prime \cite{f-prime-website}, Lightweight Communications and Marshalling (LCM) \cite{huang2010lcm}, and MOOS-IvP \cite{benjamin2010nested}}. However, these middlewares do not automatically redirect tasks among the robots based on their availability, connectivity, and state; instead, they rely on the robots' executives to select available computation resources. Also, crucially, many of these middlewares require continuous connectivity between robots, where communication disruptions might make cooperative behaviors unreliable or non-functional. In contrast, the proposed PDRA seamlessly handles intermittent connectivity when paired with a suitable transport layer (e.g., DTN \cite{burleigh2003dtn}).

\paragraph{Multi-robot task allocation}
\rev{At its core, the proposed approach relies on a task allocation mechanism that maps computing tasks to robots.  A number of task allocation approaches for networked vehicles have been proposed in the literature, including  distributed constraint optimization problems (DCOP) \cite{fioretto2018dcop,li2014mabp}, auction-based algorithms \cite{gerkey2002sold, gerkey2004formal}, hybrid planners based on second-order cone programming \cite{FernandezGonzalez2018}, and mixed-integer programming (MIP) approaches  \cite{hook-et-al-icaps19, benton2012temporal}. We refer the reader to \cite{gerkey2004formal} for an excellent survey of existing approaches to task allocation in multi-robot systems. %
Compared to these approaches, our contribution offers two key advantages, one at the algorithmic level and one at the integration level.
From an algorithmic standpoint, to the best of our knowledge, no existing distributed task allocation algorithm captures heterogeneous computation capabilities, periodically time-varying multi-hop communication links with limited throughput, task precedence constraints, and optional tasks. In particular, MIP approaches and hybrid planners are generally implemented in a centralized setting; and 
auction algorithms and DCOP algorithms have difficulty encoding capacity constraints such as availability of communication links, especially in a multi-hop communication setting. 
From an integration standpoint, existing task allocation algorithms specify \emph{where} tasks should be executed, but require a separate (and tightly integrated) planning and execution module to realize the allocation decision, interfacing 
with the communication system and
with other robots; in contrast, PDRA ``plugs in'' to existing single-agent scheduler/executives, enabling seamless integration with existing planning and execution systems.
Some existing AI planning and execution frameworks for robotics, e.g. ROSPlan~\cite{cashmore-et-al-2015-rosplan}, can be adapted to offer integration capabilities similar to PDRA; however, these frameworks are generally tailored for the setting where communication is continuously available and a centralized task allocation scheme can be used. In contrast, PDRA explicitly targets applications that present intermittent/periodic connectivity, and require distributed task allocation.
}
\subsection{Organization}
The rest of this paper is organized as follows. In Section~\ref{sec:problem-statement}, we formally define the problem that PDRA sets out to solve. In Section~\ref{sec:architecture}, we provide a detailed description of the PDRA system architecture. Section~\ref{sec:scheduling} describes the proposed MILP-based  task allocation strategy for periodically-connected networks. Section~\ref{sec:experiments} explores the performance of PDRA and of the proposed task allocation algorithm through agent-based experiments, including experiments with delay-tolerant networking and a human-in-the-loop demonstration. Finally, in Section~\ref{sec:conclusions}, we draw conclusions and discuss directions for future research.

\section{Problem Statement}
\label{sec:problem-statement}

Consider a set of mobile robots with known, time-varying communication links between them, called \textit{Contact Plan}. A contact plan specifies a set of time windows when pairs of robots can communicate, and the corresponding estimated available bandwidth. %
Each robot wishes to execute its own \textit{Software Network}, i.e., a set of mandatory and optional computational tasks whose dependencies can be represented by a directed forest graph with nodes as tasks and directed edges as dependencies.
Certain tasks in the software network can be delegated to another robot by sending the robot the inputs required for the task (which may be the output of a previous computational task); the robot performing the task should then perform the task and return the output to the robot requesting assistance.

\jvspace{-.2em}
\subsection{Assumptions}
We assume that:
\begin{itemize}
\item Computational tasks are \emph{pure functions}, i.e, the output of a computational task depends exclusively on its inputs, and has no side effects and no  associated state.
\item On each robot, an existing single-robot planning/execution process (denoted as an \emph{autonomy brain}) dispatches the robot's own computational tasks (i.e., it requests execution of the computational tasks from computational resources, described below), based on the software network, through an inter-process communication (IPC) mechanism (e.g., ROS). 
\item \rev{The software network that is dispatched by each robot, the time and energy cost required to perform a given task on a given robot, and the contact plan
are known to all robots (either statically or through a real-time consensus mechanism, discussed in Section \ref{sec:scheduling:distributed}).}
\item Each robot is able to execute a subset of the computational tasks by calling local \emph{computational resources} (e.g., function libraries, or ROS actions). Each resource can fulfill one of the tasks in the software network. %
\item A communication middleware is available to transport messages between the robots. The communication middleware can employ a store-and-forward mechanism and does not need to support acknowledgments.
\end{itemize}
\jvspace{-.2em}
\subsection{Goal}
The goal of PDRA is to provide (1) an execution engine and (2) a distributed resource allocation scheme that, together, enable task execution to be transparently shared by robotic agents. The following properties are the focus of PDRA's architecture:
\begin{itemize}
    \item \textit{Distributed}: PDRA should require no central  server and work with intermittent communication links.
    \item \textit{Optimized}: PDRA should allocate tasks the robots' computational resources so as to (approximately) optimize a given cost function, e.g., minimize energy use, or maximize the number of optional tasks completed.
    \item \textit{Pluggable}: PDRA should be integrated with existing autonomy brains, existing robotic framework (in particular, ROS), and existing computational resources (e.g., ROS packages) with minimal effort, thus maximizing code re-usability.
    \item \textit{Layered}: PDRA should enforce, to the extent possible, layering in a robot's protocol stack. In other words, PDRA should not be responsible for instantiating network transport services. Rather, PDRA should leverage existing networking layers such as TCP-IP and DTN.%
\end{itemize}

\section{System Architecture}
\label{sec:architecture}

Figure \ref{fig:pdra_simplified_architecture} presents the high-level architecture of PDRA. Each robot runs an autonomy brain and several computational resources (e.g., processes providing ROS actions). Additionally, two processes responsible for orchestrating task sharing across different robots are also instantiated. We term them the \textit{dispatcher} and the \textit{resource-obligation matcher}, or ROC. \rev{The ROC is responsible for deciding where tasks should be executed; the dispatcher ensures that tasks are correctly routed in line with the ROC's allocation. } The system also assumes the presence of a controller that provides the ROC with a consistent \emph{world view} of the system state and of tasks to be executed (discussed in Section \ref{sec:scheduling:distributed}), and a \emph{networking transport layer} that handles inter-robot communications (discussed in Section \ref{sec:architecture:chandler}). 

The unit of information of PDRA is a \textit{dispatchable} (akin to a packet in networking), which contains information about the task that must be executed and about the requesting robot. Dispatchables also contain a unique identifier and a time-to-live. Two types of dispatchables are defined: obligations and results. \emph{Obligations} encapsulate a request to perform a given computational task to be sent over the network (i.e., the identity of the requesting robot, an identifier for the computational task that should be performed, and the inputs to the computational task). \emph{Results} are created in response to an obligation once the corresponding task is executed, and contain the output of the computation and the identity of the robot that completed the task.

PDRA also defines the concept of \textit{chained obligations}, that allow a robotic agent's autonomy brain to dispatch entire computational pipelines
 as a collection of dependent obligations whose inputs are the outputs of its predecessor. E.g., consider two tasks $t_1$ and $t_2$ such that the inputs to $t_2$ are the outputs of $t_1$. Assume also that robot $R_1$ needs $t_1$ and $t_2$ performed, and its ROC indicates that $R_2$ should execute $t_1$. Then, $R_1$'s autonomy brain dispatches $t_1$ and $t_2$ to its PDRA, which encapsulates them in a chained obligation and sends them to $R_2$. $R_2$, in turn, executes $t_1$ and generates a result for $R_1$. As soon as the result of $t_1$ is computed, $R_2$'s PDRA creates a new obligation for $t_2$ and processes it through its own dispatcher and ROC, eliminating an unnecessary round-trip between $R_1$ and $R_2$. This feature is particularly useful in a DTN environment, where connectivity is dynamic and there are no guarantees as to when two robots will be in contact.

Next, we describe all elements in PDRA in detail.

\subsection{Dispatcher}
\label{sec:architecture:dispatcher}

The dispatcher is the main process for the execution engine of PDRA.
It provides a set of \emph{front-end} interfaces that intercept requests for computational tasks from the autonomy brain; a corresponding set of \emph{back-end} interfaces that forward computation requests to on-board computational resources;  one thread for a \textit{forwarder}; a second thread for an \textit{acceptor}; and a \emph{communication handler}.

\subsubsection{Dispatcher Front-End}
\label{sec:architecture:ohandler}

The dispatcher front-end is a shim layer that interfaces with the robot's autonomy brain, effectively intercepting the autonomy brain's IPC communications with computational resources.
Specifically, dispatches for computational resources from the autonomy brain are transformed to obligations and sent to the forwarder; results from the acceptor are translated to the IPC protocol understood by the autonomy brain and delivered to it. 
By transparently replacing and mimicking the IPC mechanism understood by the autonomy brain, the front-end makes PDRA \emph{pluggable}, i.e., it allows PDRA to integrated with an existing single-agent autonomy brain with minimal effort. 
In our open-source implementation, we provide a dispatcher front-end for ROS actions. Similar interfaces can be built for ROS services or any other IPC protocol. The front-end also implements a timeout system: if no result is received for a given obligation after a given amount of time, the front-end reports to the autonomy brain that the computational task has failed. This prevents processing delays or lost packets from disrupting the operations of single-agent autonomy brains designed under the assumption of reliable IPC  --- a critical consideration due to possible network delays and disruptions present in DTNs.

\subsubsection{Dispatcher Back-End}
\label{sec:architecture:rhandler}

The dispatcher back-end is another shim layer that interfaces the dispatcher with the computational resources, and its function is complementary to the front-end's.
The back-end receives obligations from the dispatcher front-end, translates them to the IPC protocol understood by the computation resources, and sends them to the resource. When a result is received from a computational resource, the back-end transforms it to a dispatchable result and sends it to the acceptor.
In our open-source implementation, we provide a dispatcher back-end for ROS actions.

\subsubsection{Dispatcher Forwarder} The forwarder is responsible for receiving dispatchables from the front-end and routing them to the appropriate back-end or to the communication handler. Both obligations and results are processed by the forwarder, albeit using different logic. For \emph{obligations}, the forwarder first determines where the obligation needs to be processed by consulting the ROC. If it determines that the obligation should be processed locally, then the forwarder sends it to the back-end. Alternatively, the obligation is routed to the communication handler. If the requested resource is not known to the ROC, the forwarder attempts to execute it locally by sending it to the back-end. If the ROC believes that a dispatchable should not be executed (e.g., an optional task can't be scheduled within the system's computational capabilities), the forwarder sends a result signaling execution failure to the acceptor.  
\emph{Results} processed by the forwarder are sent directly to the communication subsystem, as a result intended for the local robot is never processed by the forwarder (only the acceptor handles it). 

\subsubsection{Dispatcher Acceptor} The acceptor's functionality is complementary to the forwarder's. The acceptor receives dispatchables from the back-end and from the communication handler, and  determines whether they should be delivered to the autonomy brain or routed to the forwarder for further processing.
\emph{Results} arriving from local resources or from the communication handler are delivered to the front-end, which, in turn, will translate and forward them to the autonomy brain. Prior to that, however, the acceptor checks that no \emph{chained obligations} are present in the result. If that is the case, new dependent obligations are created and routed to the forwarder for further processing. 
\emph{Obligations} enter the acceptor exclusively from the communication handler. Therefore, the acceptor sends them to the forwarder which, in turn, will route them to the back-end to be executed on the applicable computational resource. Note that, prior to performing this operation, the forwarder will consult the ROC to ensure that this task should be performed locally (as it also does for obligations generated by the robot's own autonomy brain). If the ROC believes the obligation should not be performed locally, the forwarder will route the obligation back to the communication subsystem. This ensures that PDRA is resilient to robots disagreeing on the overall task allocation: e.g., if robot $R_1$ sends an obligation to robot $R_2$ for processing, but $R_2$'s ROC believes the obligation should be processed elsewhere, $R_2$ will simply forward the obligation to the robot designated by its own ROC. 
In pathological cases, obligations may be passed among robots in a cycle: to mitigate this, unfulfilled obligations are discarded after a time-to-live specified in their metadata.

\subsubsection{Communication Handler}
\label{sec:architecture:chandler}
The communication handler is the interface between PDRA and the networking transport layer that is used to transfer data across robots. The transport layer is responsible for transferring raw packets between robots; the communication handler translates packets into dispatchables and vice versa.
The communication handler can interface with a variety of inter-robot communication mechanisms including ROS channels, raw TCP sockets, DDS, and/or DTN. In our implementation, we provide communication handlers both for ROS channels (for operations on static networks) and for JPL's Interplanetary Overlay Network (ION) implementation of DTN \cite{burleigh2003dtn} (for operations on dynamic networks).

\subsection{Resource-Obligation Matcher}
\label{sec:architecture:rac}

The Resource-Obligation Matcher (ROC) allocates tasks to robots based on their computational capabilities, available network bandwidth and latency between the robots, and the software network. The interface between the ROC and the dispatcher is a dynamically updated \emph{look-up table} that specifies where each task should be executed.
In Section \ref{sec:scheduling}, we propose a MILP-based task allocation algorithm designed for networks with periodic or static connectivity that can return optimal task allocations in under a second for systems with dozens of robots. The algorithm is implemented in a distributed fashion through a shared-world approach, similar to our prior work in~\cite{hook-et-al-icaps19}.

\rev{The light coupling between the ROC and the dispatcher allows the integration of other task allocation algorithms. 
For networks with time-varying communication links that do not exhibit periodicity, we advocate for the use of the MILP-based algorithm presented in~\cite{hook-et-al-icaps19}, which captures arbitrary time-varying communication networks at the price of significantly increased computational complexity.
A variety of other distributed task allocation algorithms are also available, including auctions algorithms~\cite{gerkey2004formal} and DCOP solvers~\cite{fioretto2018dcop} --- evaluating the performance of such algorithms in concert with PDRA is an interesting direction for future research.}

\subsection{Simulation and Supervision}
\label{sec:architecture:visualization}
\ifjournalv
\begin{figure*}[h]
\centering
\includegraphics[width=.675\textwidth]{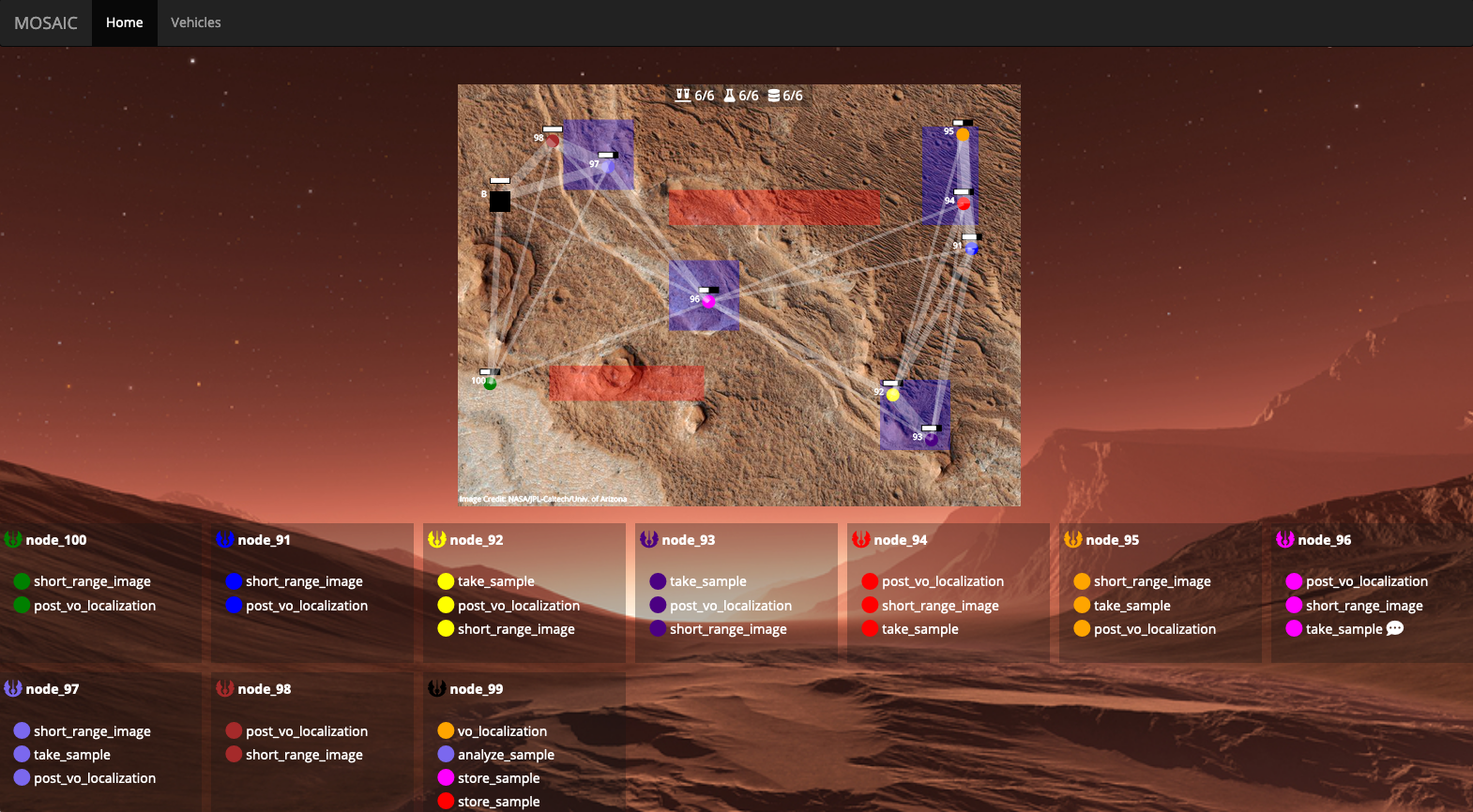}
\includegraphics[width=.2\textwidth]{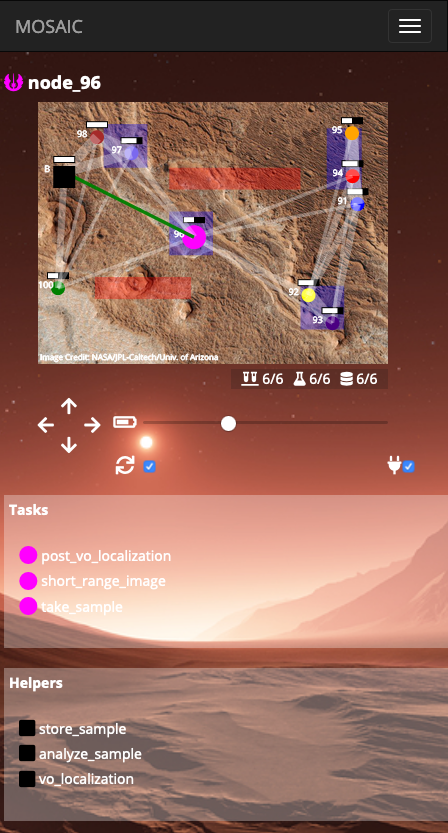}
\caption{Simulation framework and visualization for PDRA. Left: the overall system view shows robots' locations and battery levels, network link bandwidths, tasks allocated to robots, and tasks being performed. Right: the single-agent view allows manipulation of the computational capabilities and location (a proxy for network bandwidth) of individual robots.}
\label{fig:supervision}
\end{figure*}
\fi

We developed a set of agent-based simulation and supervision tools to assess the performance of PDRA. The simulation framework allows users to create and edit scenarios with a varying number of robots with heterogeneous computational capabilities, and change communication bandwidths between robots, through
\ifjournalv
the graphical user interface (GUI) shown in Figure~\ref{fig:supervision}.
\fi
 The GUI also enables real-time monitoring of task allocation and execution, and it was successfully used with non-expert users in a human-in-the-loop demonstration at the 2019 ICAPS conference, described in Section~\ref{sec:experiments:icaps}. 

\section{Allocating tasks in a mobile robotic network}
\label{sec:scheduling}

In this section, we propose a task allocation algorithm suitable for multi-robot systems with \emph{periodic} or \emph{static} network connectivity. First, we describe a \emph{centralized} implementation. Next, in Section \ref{sec:scheduling:distributed}, we propose a \emph{distributed} implementation through a shared-world approach.

The goal of the algorithm is to allocate computational tasks to robots while satisfying constraints on:
\begin{itemize}
\item[(i)] Logical dependencies between tasks: certain tasks require as input the outputs of other tasks;
\item[(ii)] Maximum latency of data products: the inputs to a task must be delivered to the performing robot within a maximum latency from the moment they are computed;
\item[(iii)] Maximum computational load of computing robots;
\item[(iv)] Maximum throughput of communication links.
\end{itemize} 

We cast this problem as a mixed-integer linear program (MILP) in a network flow framework \cite{AhujaMagnantiEtAl1993}.
The key modeling assumption of the task allocation algorithm is that the average available throughput on communication links is approximately constant over a representative time period $\timescale$.
With this assumption, the problem is modeled through a time-invariant graph, where nodes represent robots, edges represent communication links, and a set of binary decision variables capture the allocation of tasks to robots.
Dispatchables transmitted from a robot to another are modeled as a network flow with an origin at the robot dispatching the task to be executed (or, in the case of chained obligations, performing the parent task) and a destination at the robot actually executing the task.

\subsection{Problem Parameters}
\label{sec:scheduling:parameters}
The problem is parameterized by the following properties:
\begin{itemize}
\item $\timescale$, the time period of interest;
\item $\tasks$, the set of computational tasks to be allocated in the time period of interest;
\item $\taskchildren{\task}$, the set of tasks that require task $t$'s data products as inputs. Tasks $\tau\in \taskchildren{\task}$ are denoted as \emph{children} of $\task$; 
\item $\taskproductsize{\task}$, the size of the data product of task $t$, in bits;
\item $\requiredtasks \subseteq \tasks$, the set of \emph{required} tasks that must be executed;
\item $\optionaltasks = \tasks \setminus \requiredtasks$, the set of \emph{optional} tasks;
\item $\reward{\task}$, the reward for planning the execution of optional tasks $\task\in\optionaltasks$;
\item $\agents$, the set of available robots;
\item $\links$, the set of pairs of robots $(i, j)$ that can directly communicate through a radio link at some point during the time period of interest;
\item $\taskcomputationload{\task}{\agent}$, the computational load (i.e., the average fraction of a CPU core, averaged over the time period \timescale) required to perform task $t$ once on robot $i$;
\item $\taskpower{\task}{\agent}$, the average power required to perform task $t$ once on robot $i$ over time \timescale, in Watts;
\item $\linkcomputationloadout{i}{j}$, the computational load (i.e., the fraction of a CPU core) required to encode one bit of information per second on the outbound stream on link $(i,j)$;
\item $\linkcomputationloadin{i}{j}$, the computational load to decode one bit of information per second from the inbound stream on $(i,j)$;
\item $\linkenergyout{i}{j}$, the energy (in J/bit) required to encode and transmit one bit per second on the outbound stream on link  $(i,j)$;
\item $\linkenergyin{i}{j}$, the energy (in J/bit) required to decode one bit per second from the inbound stream on link $(i,j)$;
\item $\maxthroughput{i}{j}$, the maximum available throughput (in bits/s) on the radio link $(i,j) \in \links$, averaged over the time period of interest;
\item $\lslatency{i}{j}$, the light-speed latency of link $(i,j)$ (in s);
\item $\maxcomputationload{\agent}$, the maximum computational load of robot $i$ (i.e., the number of CPU cores available at robot $\agent$);
\item $\maxlatency{\task}{\tau}$ is the maximum acceptable latency (in s) for the data products of task $\task$ required by task $\tau$.
\end{itemize}

\subsection{Optimization Variables}
The optimization variables are:
\begin{itemize}
\item $X$, a matrix of Boolean variables of size $|\agents|$ by $|\tasks|$. $X(i,t)$ is set to 1 if task $t$ is assigned to robot $i$ and $0$ otherwise;
\item $C$, a matrix of real-valued variables of size $|\links|$ by $\sum_{t\in\tasks}|\taskchildren{t}|$. $C(i,j,t,\tau)$ is the amount of data products of task $t$ (in bits per second) transmitted on link $(i,j)$ and that will be used as input by task $\tau$;
\item $B$, a matrix of real-valued variables of size $|\links|$ by $|\tasks|$. $B(i,j,t)$ is the amount of bandwidth used by data products of task $t$ on link $(i,j)$ (irrespective of the destination). 
\end{itemize}

The variables $C$ keep track of data products separately according to the requiring child task; this enables the proposed model to (1) capture the average latency of the data products to multiple destinations (as discussed in Section \ref{sec:MILP:latency}) and (2) represent logical dependencies between tasks by considering the requiring task as an ``information sink''. However, identical data transmitted on the same link should not be duplicated, irrespective of the destination. Variables $B$ capture the fact that, in a practical implementation, different data flows $C$ containing the same data product are de-duplicated and transmitted once per communication link.

\subsection{Optimization Objective}

The optimization objective is the weighted sum of two terms capturing (1) the reward for completing optional tasks and (2) the power required to solve the problem.
The individual cost functions can be captured as follows.
\begin{subequations}
\begin{itemize}
\item Maximize the sum of the rewards for completed optional tasks:
\begin{equation}
R_r =   \sum_{j\in\agents} \sum_{t\in\tasks}  \reward{t} X(j,t) \label{eq:MILPgoal}
\end{equation}
\item Minimize the power cost of the problem:
\begin{equation}
R_e = - \sum_{j\in\agents} \sum_{t\in\tasks} \taskpower{t}{j}X(j,t) + \sum_{(i,j)\in\links} \sum_{t\in\agents} (\linkenergyout{i}{j}+\linkenergyin{i}{j}) B(i,j,t)
\end{equation}
\end{itemize}

The optimization objective can then be expressed as
$
R = \alpha R_r + (1-\alpha) R_e,
$
where $\alpha$ is the relative weight of the two reward functions.
\label{eq:MILP:costs}
\end{subequations}

\subsection{Problem Formulation}
The problem can be posed as a mixed-integer linear program as follows.
\jvspace{-2mm}

\begin{subequations}\label{eq:MILP}
\begin{align}
&\max_{X, C, B}  R \label{eq:MILP:cost}\\
&\text{subject to:}\nonumber \\
& \sum_{j\in \agents} X(j,t) = 1 \qquad \forall t \in \requiredtasks \label{eq:MILP:requiredtasks} \\
& \sum_{j\in \agents} X(j,t) \leq 1 \qquad \forall t \in \optionaltasks \label{eq:MILP:optionaltasks} \\
& X(j,t) \frac{\taskproductsize{t}}{\timescale} + \sum_{i: (i,j)\in\links} C(i,j,t,\tau) \geq X(j,\tau) \frac{\taskproductsize{t}}{\timescale} +  \nonumber \\
&\quad \sum_{k:(j,k)\in\links} C(j,k,t,\tau) \quad \forall j \in \agents, t\in\tasks, \tau \in \taskchildren{t} \label{eq:MILP:prereqs} \\
& B(i,j,t) \geq C(i,j,t,\tau) \quad \forall (i,j) \in \links, t\in \tasks, \tau\in \taskchildren{t} \label{eq:MILP:multiplex}\\
& \sum_{t\in\tasks } \taskcomputationload{t}{j} X(j,t) + \sum_{i: (i,j) \in\links}\sum_{t\in\tasks} \linkcomputationloadin{i}{j}B(i,j,t) \nonumber \\
& \quad + \sum_{k: (j,k) \in\links}\sum_{t\in\tasks} \linkcomputationloadout{j}{k} B(j,k,t) \leq \maxcomputationload{j}  \qquad \forall j\in\agents \label{eq:MILP:computation}\\
&\sum_{t\in\ \tasks } B(i,j,t) \leq \maxthroughput{i}{j} \qquad \forall (i,j) \in \links \label{eq:MILP:bandwidth}\\ %
& \sum_{(i, j) \in\links} \left(\lslatency{i}{j} + \frac{\taskproductsize{t}}{\maxthroughput{i}{j}}\right) C(i,j,t, \tau) \frac{\timescale}{\taskproductsize{t}}    \leq \maxlatency{t}{\tau} \nonumber \\
& \qquad   \forall t\in \tasks, \tau   \in \taskchildren{t}
\label{eq:MILP:latency}
\end{align}
\end{subequations}

Equation~\eqref{eq:MILP:requiredtasks} ensures that all required tasks are allocated. Equation~\eqref{eq:MILP:optionaltasks} ensures that optional tasks are allocated at most once. Equation~\eqref{eq:MILP:prereqs} (discussed in Section~\eqref{sec:MILP:prereqs}) ensures that robots only execute a task if they have access to the data products of its pre-requisites.
Equation~\eqref{eq:MILP:multiplex} ensures that the variable $B$ is an upper bound on the data throughput used to transmit data products of task $t$ on each link.
 Equation~\eqref{eq:MILP:computation} enforces that the computational capacity of each robot is not exceeded. Equation~\eqref{eq:MILP:bandwidth} ensures that the transmissions on a link do not exceed the available bandwidth. Finally, Equation~\eqref{eq:MILP:latency} (discussed in Section~\ref{sec:MILP:latency}) guarantees that data products are delivered to dependent tasks within a maximum acceptable latency. 

\subsubsection{Prerequisites}\label{sec:MILP:prereqs}

Equation~\eqref{eq:MILP:prereqs} ensures that a copy of the data product of task $t$ is created at the robot where task $t$ is performed, and it is consumed by the robot where task $\tau$ is performed. At other robots, information can be relayed or deleted, but not created.

\subsubsection{Latency}\label{sec:MILP:latency}
Equation~\eqref{eq:MILP:latency} ensures that the \emph{average} latency of data products of task $t$ used by task $\tau$ is smaller than $\overline{l}_{t,\tau}$. The latency for the data products of task $t$ destined for task $\tau$ on a \emph{given} path $p=\{(u,v), (v, w), \ldots, \}$ is $\sum_{(i,j) \in p} (\lslatency{i}{j} + \taskproductsize{t}/\maxthroughput{i}{j})$, where \taskproductsize{t} is the size (in bits) of the output of task $t$, and $\lslatency{i}{j}$ and $\maxthroughput{i}{j}$ are respectively the light-speed latency and the maximum available bandwidth on link $ij$. If data products are sent on multiple paths $p_1, \ldots, p_m$, with path $p_k$ carrying a fraction $f_k$ of the data product, the average latency is:
\begin{align*}
 l_{t,\tau} & = \sum_{k=1}^m f_k \sum_{(i,j)\in p_i} (\lslatency{i}{j} + \taskproductsize{t}/\maxthroughput{i}{j}) \\
 & = \sum_{(i,j) \in \links} (\lslatency{i}{j} + \taskproductsize{t}/\maxthroughput{i}{j}) \sum_{k=1}^m 1_{(i,j)\in p_k} f_k
\end{align*}
The flow $C(i,j,t,\tau)$ captures the sum of path flows on all paths carrying the data products of $t$ for task $\tau$. The overall amount of flow on all paths is the size of the data product $d_t$ divided by the time period \timescale. Accordingly, the average latency can be rewritten as
\[
l_{t,\tau} = \sum_{(i,j) \in \links} (\lslatency{i}{j} + \taskproductsize{t}/\maxthroughput{i}{j}) C(i,j,t,\tau)\frac{\timescale}{\taskproductsize{t}}
\]
in accordance with Equation~\eqref{eq:MILP:latency}.

\subsection{Distributed Implementation}
\label{sec:scheduling:distributed}
Solving Problem \eqref{eq:MILP} requires knowledge of \emph{global} information from all robots, namely the network bandwidths, the robot's computational capabilities, the software network, and the task computational loads, power requirements, and rewards.
In order to achieve a truly distributed resource allocator, we advocate for a \emph{shared-world} approach where robots share global information with each other through a message-passing protocol (e.g., gossip or flooding~\cite{lynch1996distributed}). Each robot maintains a \emph{world view} containing an up-to-date representation of the global system state
(namely, a compact representation of the parameters described in Section \ref{sec:scheduling:parameters}),
 and solves Problem \eqref{eq:MILP} based on its world view.

For many applications, only a moderate  amount of communication is required to build a world view at execution time. If the type of robots that will participate in PDRA is known before deployment, the robots' computational capabilities can be parametrized by a small number of bits (corresponding, e.g., to battery level). The software network can also be generally represented in a compact fashion; for instance, robots executing the software network shown in Figure~\ref{fig:scenario_task_network} only need to transmit one bit of information to inform others of whether they want to allocate tasks in the ``science'' task chain. Finally, bandwidth on radio links is typically negotiated by the communication protocol among a limited number of options; for instance, the 802.11g WiFi specification allows up to eight possible bandwidths, and the Zigbee specification only provides a single bandwidth.

Nevertheless, networking disruptions can cause the robots' world views to temporarily disagree, which may result in inconsistent task allocations. The design of PDRA is relatively robust to such inconsistencies: whenever a robot receives an obligation that its ROC believes should be processed by another robot, the obligation is simply forwarded to the robot designated by the local ROC. Pathological cases where obligations are forwarded in a cycle are prevented by endowing obligations with a maximum time-to-live.  \rev{We remark that PDRA does \emph{not} guarantee that tasks will be complete in presence of arbitrary network failures; however, the design does ensure that (1) the system will recover from short-lived disagreements in the world-views, by re-evaluating each tasks's allocation immediately before starting execution, and (2) computational tasks are guaranteed to either succeed or fail within a pre-defined amount of time, thanks to the provided time-out mechanism.}

\subsection{Discussion}

A few comments are in order.
The proposed task allocation algorithm is designed for networks with time-invariant or periodic connectivity (e.g., scenarios where an orbiter periodically overflies rovers, or where a vehicle periodically travels between clusters of robots). In particular, the computational loads, bandwidths, and energy costs can be interpreted as time-averaged quantities over the time period \timescale. When a solution to the task allocation problem is executed with the PDRA, tasks are dispatched and messages are exchanged in order of arrival; while robots' computation resources and communication links may be temporarily oversubscribed, resulting in buffering, if the time period \timescale~is chosen to be a multiple of the communication network's period, the average CPU usage and network throughput will match those prescribed by Problem~\eqref{eq:MILP}. 
Numerical experiments in~\ref{sec:experiments:numerical} show that PDRA and the ROC perform well over a network with periodic connectivity equipped with a delay-tolerant networking layer, employing a traveling robot as a ``data mule'' to share computational loads across robots that can never communicate instantly.
We emphasize that the proposed algorithm is not suitable for networks with time-varying, non-periodic connectivity. In these scenarios, we advocate for the use of the scheduling algorithm proposed in our prior work~\cite{hook-et-al-icaps19}, which explicitly captures the network's time-varying nature at the price of higher computational cost.

\section{Experiments}
\label{sec:experiments}
\subsection{Experiment Setup}
\ifjournalv
\begin{figure*}
  \centering
  \includegraphics[width=.7\textwidth]{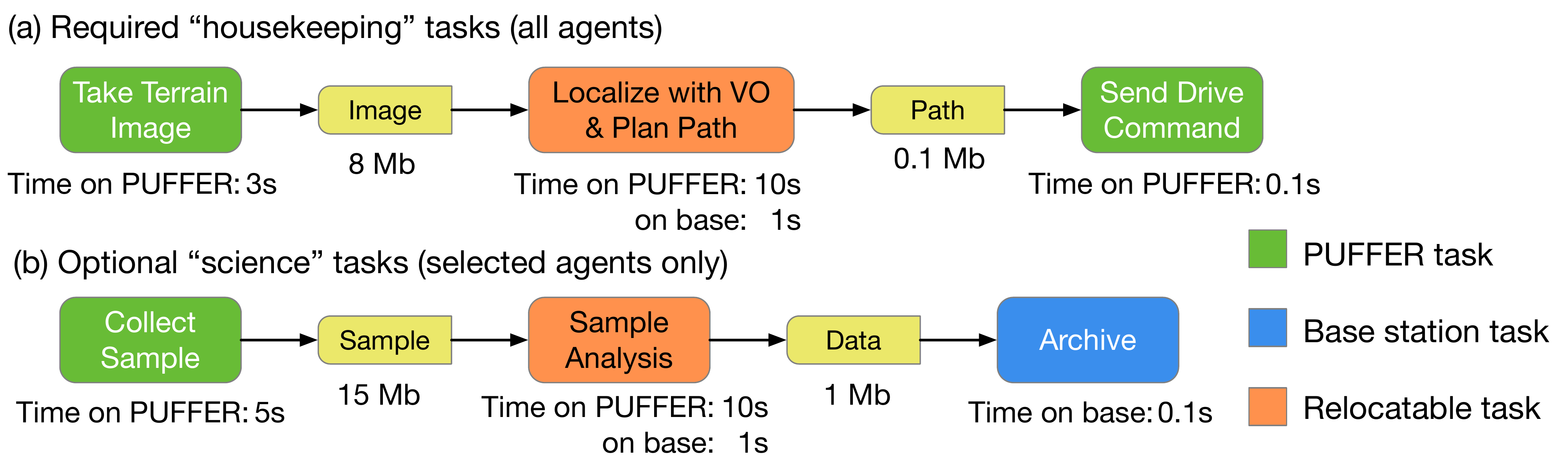}
  \caption{Software network used in numerical experiments. All robots must perform ``housekeeping'' tasks in task chain (a). Selected robots can also perform optional tasks in the ``science'' task chain (b). Computational tasks that can be relocated to another robot are represented in orange.
  }
  \label{fig:scenario_task_network}
\jvspace{-.1em}
\end{figure*}
\fi
We assessed the performance of the PDRA and of the proposed task allocation algorithm on a variety of scenarios where robots with limited computation capabilities (inspired by JPL's PUFFER robot \cite{karras2017puffer}) execute the software network in Figure~\ref{fig:scenario_task_network}. 
All robots must perform the ``housekeeping'' task chain (Figure~\ref{fig:scenario_task_network}a). Selected robots can also perform optional tasks in the ``science'' task chain (Figure~\ref{fig:scenario_task_network}b). Robots are aided by a ``base station'' equipped with a powerful and efficient CPU; the base station's computational resources must be shared among all robots. Bandwidths between robots are determined based on their locations and on a simple distance-based threshold model.%

\subsection{Task Allocation Performance}
\label{sec:experiments:scheduler}

First, we assessed the performance of the proposed task allocation algorithm in isolation by solving randomly-generated instances of Problem \eqref{eq:MILP} with an increasing number of robots. %
We consider twenty scenarios with randomly-selected robot locations\ifjournalv ~(shown in Figure \ref{fig:experiments:instances}).\fi~For each scenario, we increase the number of active robots from 2 to 16 to assess the scalability of the proposed approach.
Bandwidths are computed according to inter-robot distance.%
Each robot executed either both task chains in Figure \ref{fig:scenario_task_network} (with probability 0.6), or only the task chain in Figure \ref{fig:scenario_task_network}a (with probability 0.4).  For each instance, we consider three cost functions, namely (1) maximizing the reward for optional tasks $R_r$, (2) minimize power consumption $R_e$, and (3) a hybrid reward $0.5 R_r+0.5 R_e$.

\ifjournalv
\begin{figure}
\includegraphics[width=\columnwidth]{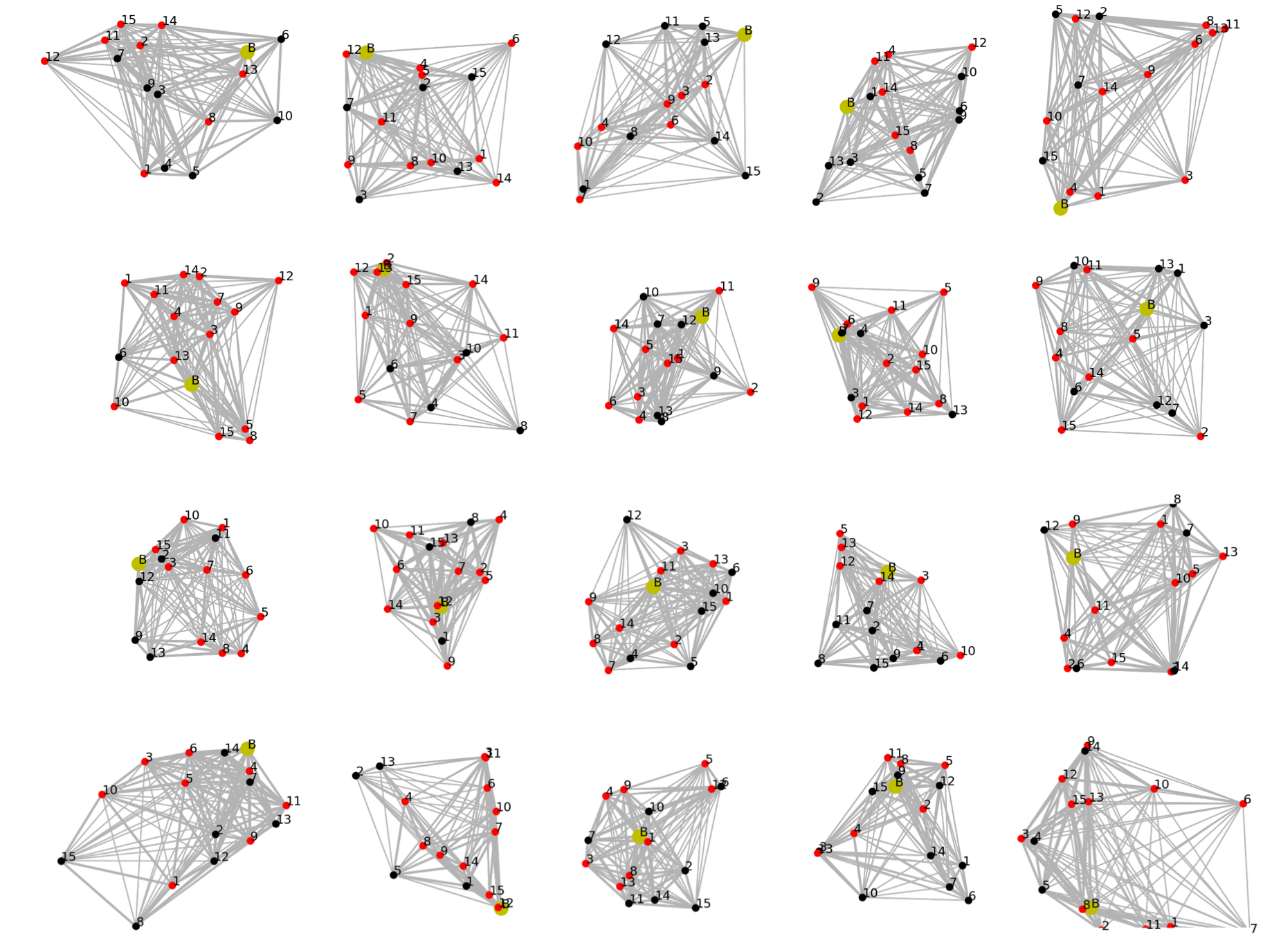}
\caption{Twenty scenarios considered in the numerical experiments. The base station is shown in yellow. Nodes able to perform science tasks are shown in red; nodes unable to perform science tasks are shown in black.}
\label{fig:experiments:instances}
\end{figure}
\fi

\begin{figure}
\begin{subfigure}[h]{\ifjournalv .85 \fi\columnwidth}
\centering
\includegraphics[width=\columnwidth]{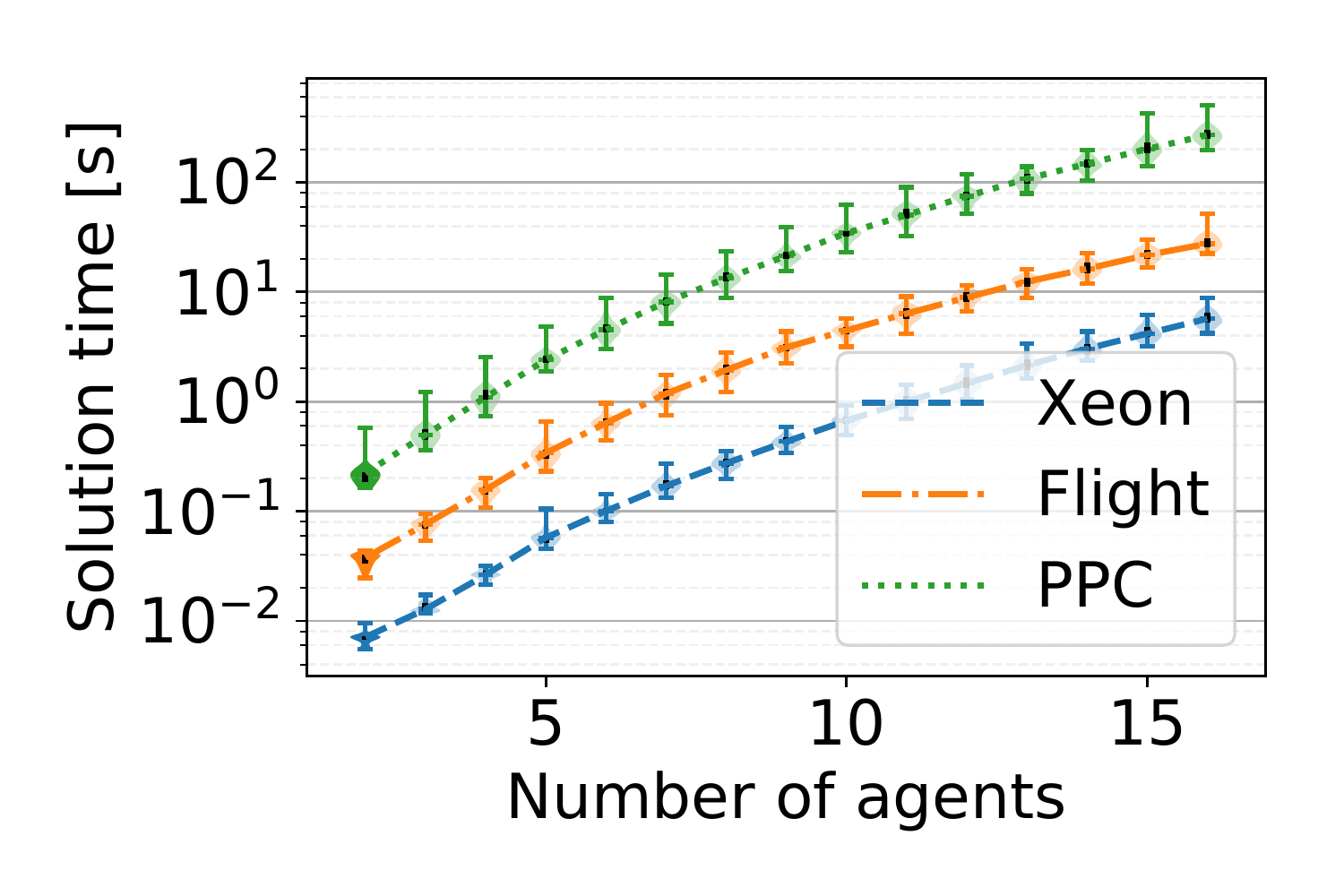}
\label{fig:experiments:benchmark:time}
\end{subfigure}
\begin{subfigure}[h]{\ifjournalv .85 \fi\columnwidth}
\centering
\includegraphics[width=\columnwidth]{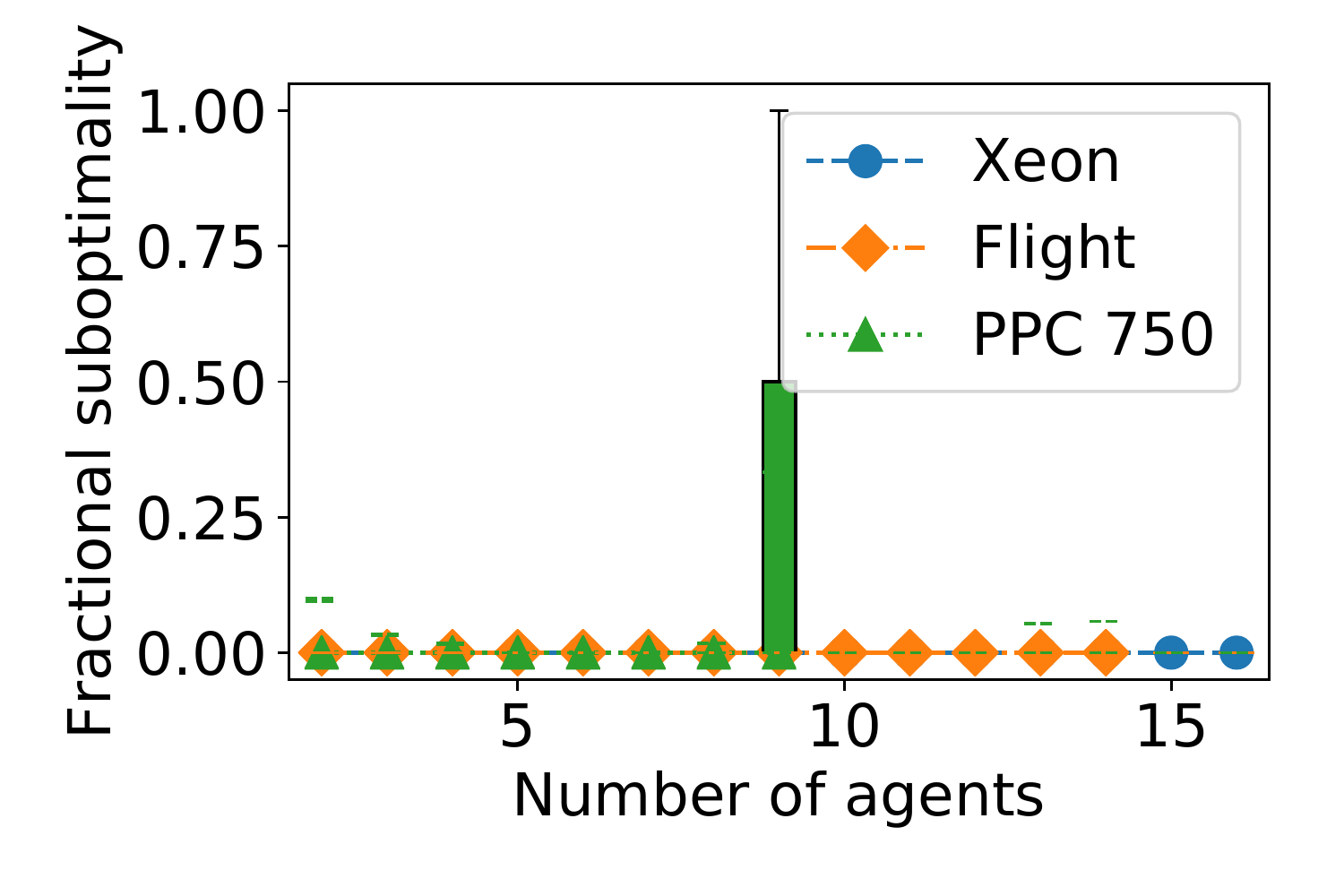}
\label{fig:experiments:benchmark:quality}
\end{subfigure}
\jvspace{-.75em}
\caption{\rev{\emph{\ifjournalv Top\fi}: Time required to solve Problem \ref{eq:MILP} as a function of number of robots. \emph{\ifjournalv Bottom\fi}: suboptimality gap of best solution found after 10s of execution, normalized by the optimal solution value.}}
\label{fig:experiments:benchmark}
\end{figure}

The problem was solved on several computing platforms, specifically:
\begin{itemize}
\item a modern Intel Xeon workstation equipped with a 10-core E5-2687W processor;
\item an embedded Qualcomm Flight platform equipped with a APQ8096 system-on-a-chip;
\item a Powerbook G3 equipped with a single-core PowerPC 750 clocked at 500 MHz, the same processor (albeit with higher clock speed) as the RAD750 used on the Curiosity and Mars 2020 rovers~\cite{Bajracharya2008autonomy}.
\end{itemize}
The MILP was solved with the SCIP solver~\cite{GleixnerEtal2018OO}. For each problem, we computed the time required for the solver to find and certify an optimal solution, and the quality of the best solution obtained after 10 seconds of execution. Results are shown in Figure~\ref{fig:experiments:benchmark}. %

On the modern Xeon architecture, the proposed task allocation algorithm is able to find and certify the optimal solution to problems with up to 11 robots in under 1\si{\second}, and it always finds the optimal solution within 10\si{\second}. The embedded Qualcomm SoC is also able to solve problems with up to 6 robots to optimality in under 1\si{\second}, and it returns the optimal solution to problems with up to 12 robots in under 10\si{\second}. Finally, even the highly limited PPC 750 processor is able to compute the optimal solution to problems with 7 robots in under 10\si{\second} -- a remarkable achievement for a 20-year-old computing architecture.

\subsection{Agent-Based Simulations}
\label{sec:experiments:numerical}

Next, we assessed the end-to-end performance of the PDRA, paired with the proposed task allocation algorithm, through agent-based simulations \rev{in two representative scenarios}. The simulation setup is shown in Figure~\ref{fig:experiments} and uses the software network in Figure~\ref{fig:scenario_task_network}. Nine low-power robots were required to perform housekeeping tasks; robots in the dark shaded regions, denoted as "science regions", were also able (but not required) to collect, analyze, and store a sample. The robots were aided by a base station.

\ifjournalv
\begin{figure}[h]
\centering
\begin{subfigure}[h]{\ifjournalv.48\else.32\fi\textwidth}
\includegraphics[width=\columnwidth]{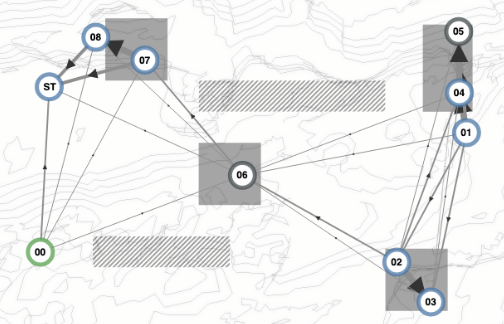}
\caption{Stationary scenario}
\label{fig:experiments:setup}
\end{subfigure}
\begin{subfigure}[h]{\ifjournalv.48\else.32\fi\textwidth}
\includegraphics[width=\columnwidth]{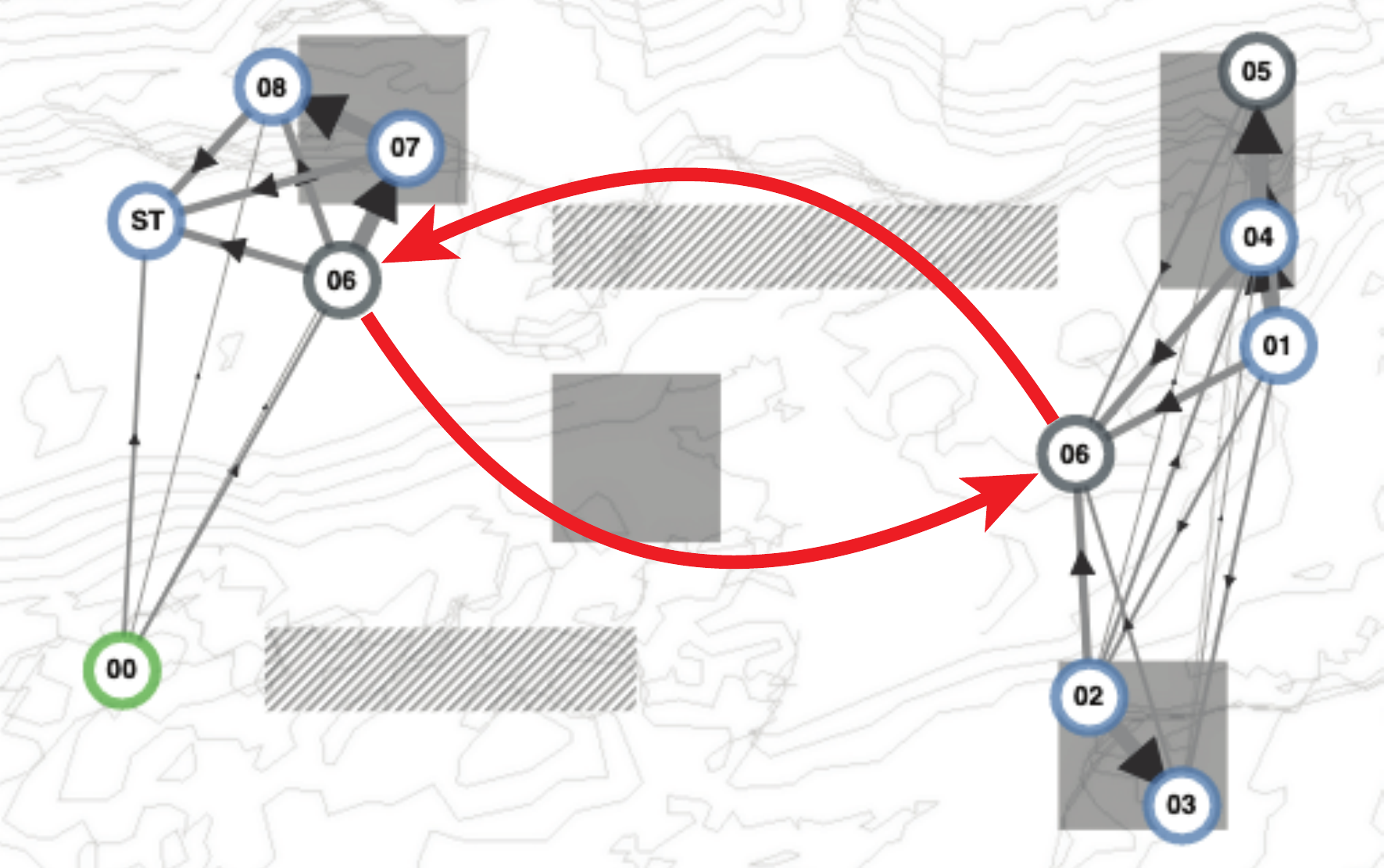}
\caption{Delay-tolerant networking (DTN) scenario}
\label{fig:experiments:dtn:setting}
\end{subfigure}
\caption{Agent-based simulations setup. Ten robots, including a base station, are placed in a 2D environment. Robots have different battery levels, which affect execution speed and energy cost of tasks. Bandwidths are based on inter-robot distance; no communication is possible through hatched regions. Robots in dark shaded regions can collect, analyze, and store samples.
In the stationary scenario (top), robots do not move. In the delay-tolerant scenario (bottom), robot 6 periodically travels between the two positions highlighted by the red arrows with a 1\si{\minute} period. At no time does a direct link exist between the robots on the left-hand side and on the right-hand side of the map. 
\jvspace{-.75em}
}
\label{fig:experiments}
\end{figure}
\fi

\rev{In the simulation, each robot maintains an independent world view containing the values of all parameters listed in Section \ref{sec:scheduling:parameters} (namely, the software network, the communication bandwidths, the agents' computational capabilities, and the power and energy costs for computation and communication), and independently solves Problem \eqref{eq:MILP}. Robots exchange messages reporting (1) the available bandwidths on communication links they participate in, (2) whether they are in a "science region", and (3) their own battery level, which affects computational capabilities. Each robot updates its world view based on other robots' messages and agreed-upon rules (e.g., robots in a science region attempt to schedule optional tasks, and a low battery level corresponds to an increase in processing times).}

We considered two scenarios: a \emph{time-invariant} scenario where the network topology is fixed (Figure \ref{fig:experiments:setup}), and a \emph{delay-tolerant networking} scenario where a "data mule" robot periodically travels between two disconnected groups of robots (Figure \ref{fig:experiments:dtn:setting}). 
In the delay-tolerant scenario, we set the planning time period $\timescale$ to equal one full cycle of the traveling robot, i.e., $60\si{\second}$.

We compared the proposed approach with
(1) a \emph{selfish} approach where Problem~\eqref{eq:MILP} is used to determine which optional tasks should be performed and which optional tasks should be skipped, but robots are not allowed to share computational tasks (except for storing samples, which is performed by the base station), and 
(2) a \emph{naive} approach where each robot performs \emph{all} available tasks on board (except for storing samples).

\begin{figure}[h]
\centering
\includegraphics[width=\columnwidth]{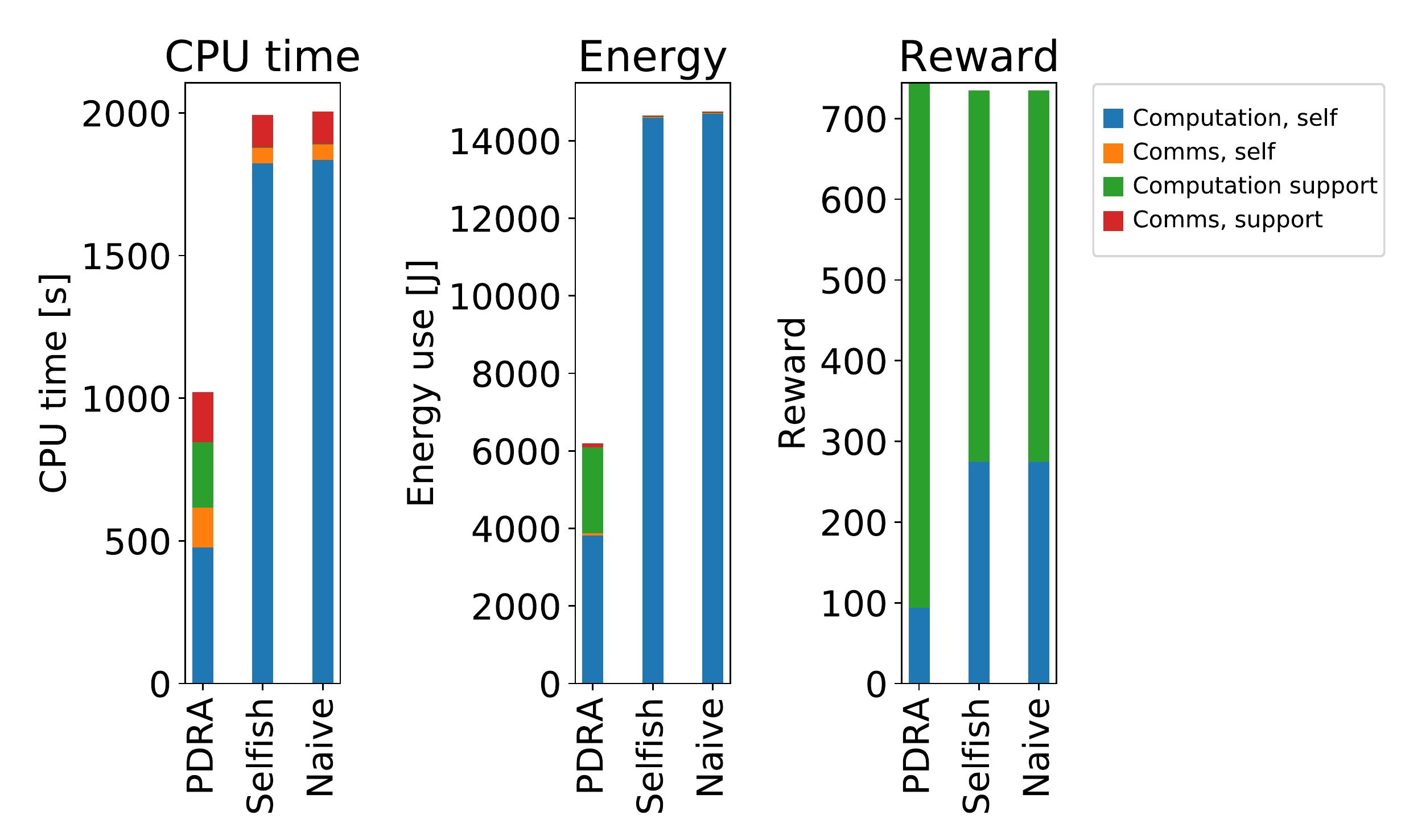}
\caption{Time-invariant agent-based simulations: CPU usage, energy use, and reward for optional tasks obtained by all agents. We compare PDRA, a naive task allocation algorithm that executes all available tasks on board, and a ``selfish'' version of PDRA that can skip optional tasks but not use computational assists.}
\label{fig:experiments:pdra:stats}
\end{figure}

\ifjournalv
\begin{figure}[h]
\includegraphics[width=.48\textwidth]{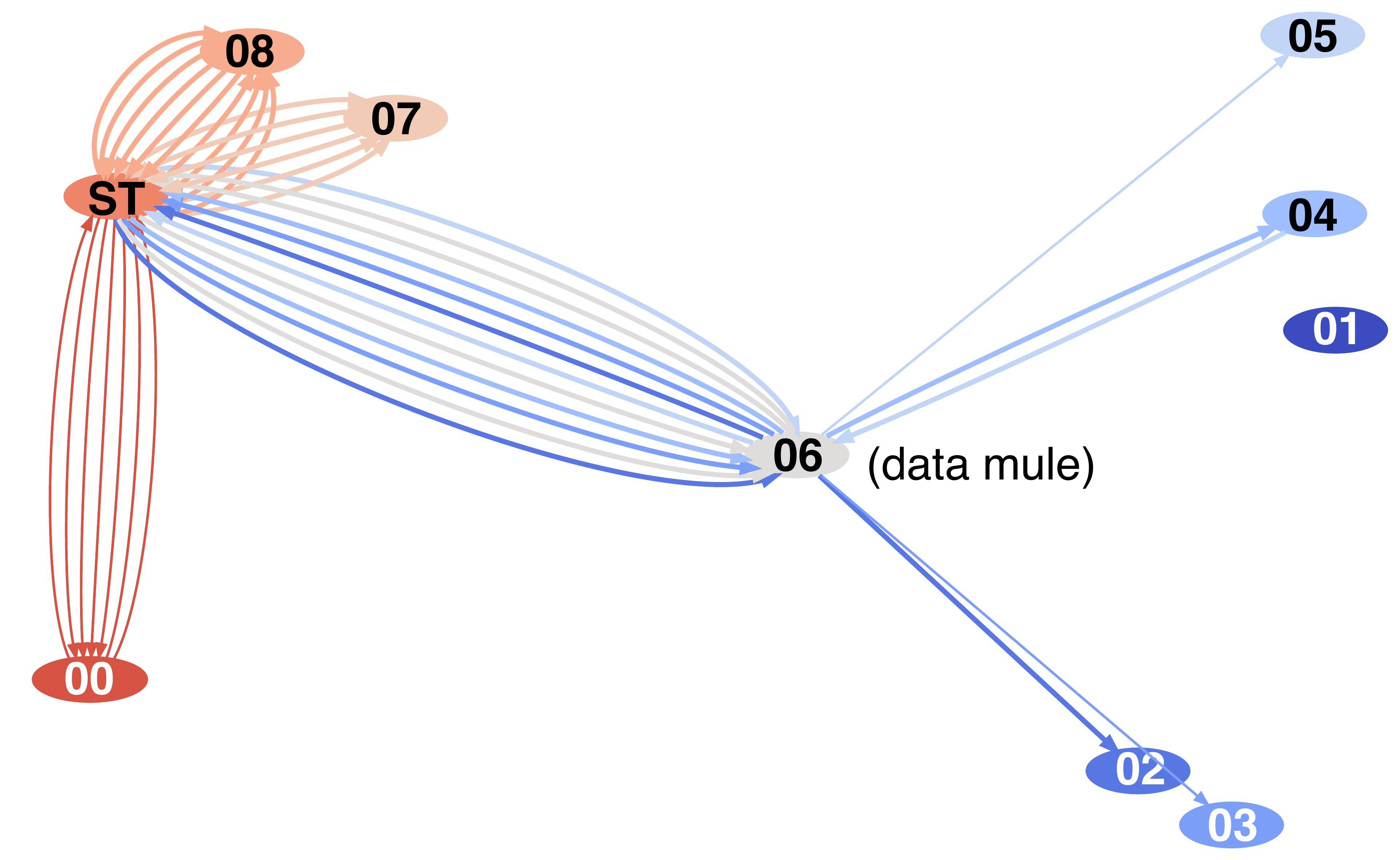}
\caption{Data flows in the DTN agent-based simulations.  Robots in the left cluster are colored in red, and robots in the right cluster are colored in blue. Edges denote messages exchanged between robots, and edge color corresponds to the robot whose data is transmitted. PDRA is able to use the moving robot as a data-mule, enabling robots on the right-hand side of the graph to offload computational tasks to the base station (labeled ST) on the left, despite the lack of an instantaneous communication link.}
\label{fig:experiments:dtn:dataflows}
\end{figure}
\fi

\rev{For each scenario, we show the results of one representative simulation.}
Results are reported in Table~\ref{tab:experiments:pdra} and Figures~\ref{fig:experiments:dtn:dataflows} and~\ref{fig:experiments:pdra:stats}. 
In the time-invariant scenario, the use of PDRA results in a 58\% reduction in system-wide energy use and a nearly 50\% reduction in overall CPU time compared to both the naive and the selfish approaches, together with a small increase in the reward for completed tasks. 
In the DTN scenario, Problem \eqref{eq:MILP} is able to exploit the moving robot as a \emph{data mule}: as shown in Figure \ref{fig:experiments:dtn:dataflows}, robots in one cluster offload computational tasks to robots belonging to the other cluster, even though no instantaneous communication link is available. PDRA routes obligations over the DTN networking layer, realizing the requested task allocation.
 
Collectively, these results show that PDRA can help realize the promise of distributed computing in heterogeneous robotic networks, greatly reducing CPU usage and energy costs while increasing science returns, with minimal changes to existing planning/execution modules and existing computational resources.

\begin{table}[h]
\jvspace{-4mm}
\centering
\begin{tabular}{cr|cc|ccc}
&\multicolumn{1}{c}{} & &  \multicolumn{1}{c}{}& \multicolumn{3}{c}{Samples} \\
&& CPU [\si{\second}] & Energy [J] & Taken & Analyzed & Stored  \\
\hline
\multirow{3}{*}{\rotatebox{90}{TI}}&PDRA  & \textbf{1022}  & \textbf{6042}& 24 & 24 & \textbf{24}  \\
&Selfish  & 1992 & 14654 & \textbf{25} & \textbf{25} & 23  \\
&Naive  & 2004 & 14750  & \textbf{25} & \textbf{25} & 23 \\
\hline
\multirow{3}{*}{\rotatebox{90}{DTN}}&PDRA  & \textbf{1531}  & \textbf{10545}& \textbf{19} & \textbf{18} & \textbf{17}  \\
&Selfish  & 1892 & 14217 & 18 & 15 & 15  \\
&Naive  & 1957 & 14748  & 18 & 17 & \textbf{17} \\
\end{tabular}
\caption{ \small PDRA performance over a 6-minute agent-based simulation. Compared to both the naive and the selfish task allocation algorithms, PDRA is able to reduce energy use by 58\% and CPU usage by almost 50\%, even accounting for communications. }
\label{tab:experiments:pdra}
\end{table}
\jvspace{-4mm}

\subsection{Human-in-the-loop demonstration at ICAPS}
\label{sec:experiments:icaps}
\begin{figure}[h]
\centering
\begin{subfigure}[h]{.85\columnwidth}
\centering
\includegraphics[width=\ifjournalv .95 \fi \columnwidth]{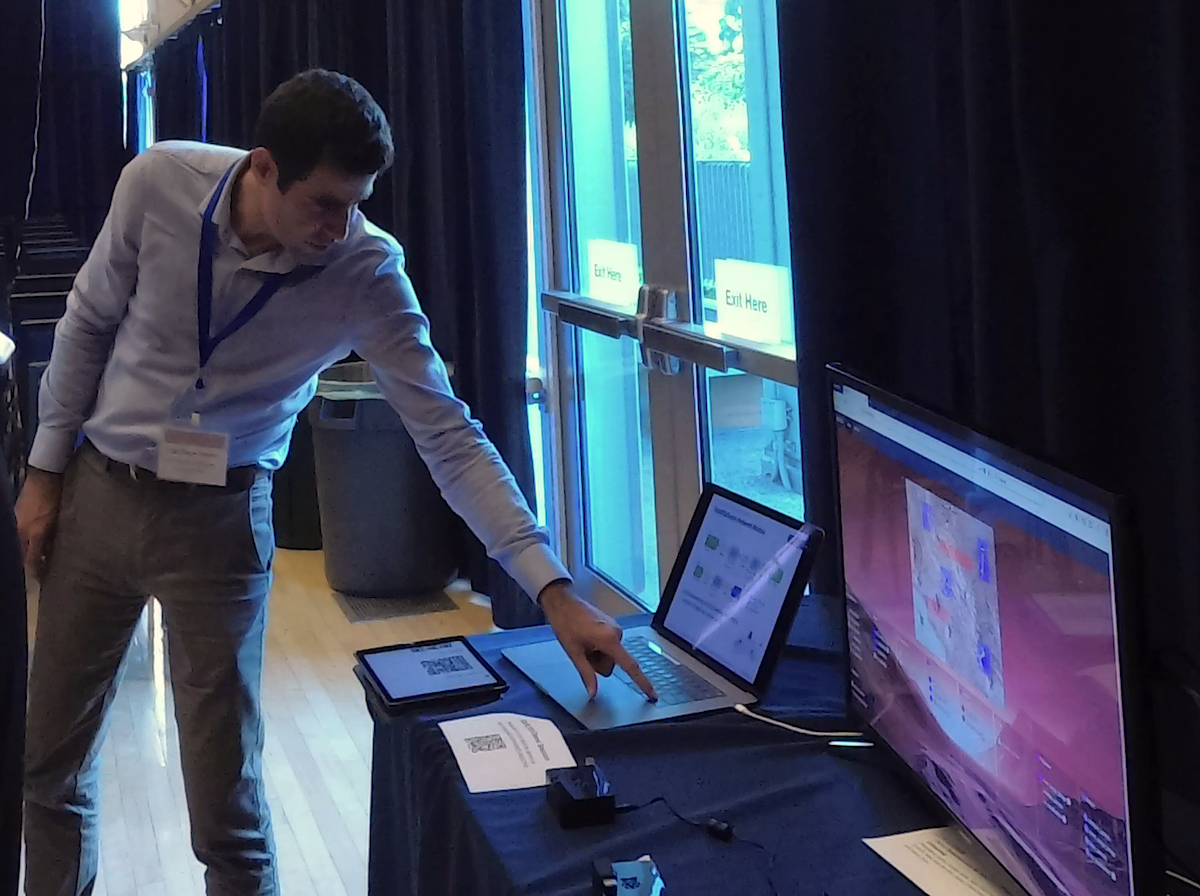}
\end{subfigure}
\begin{subfigure}[h]{.97\columnwidth}
\centering
\includegraphics[width=\columnwidth]{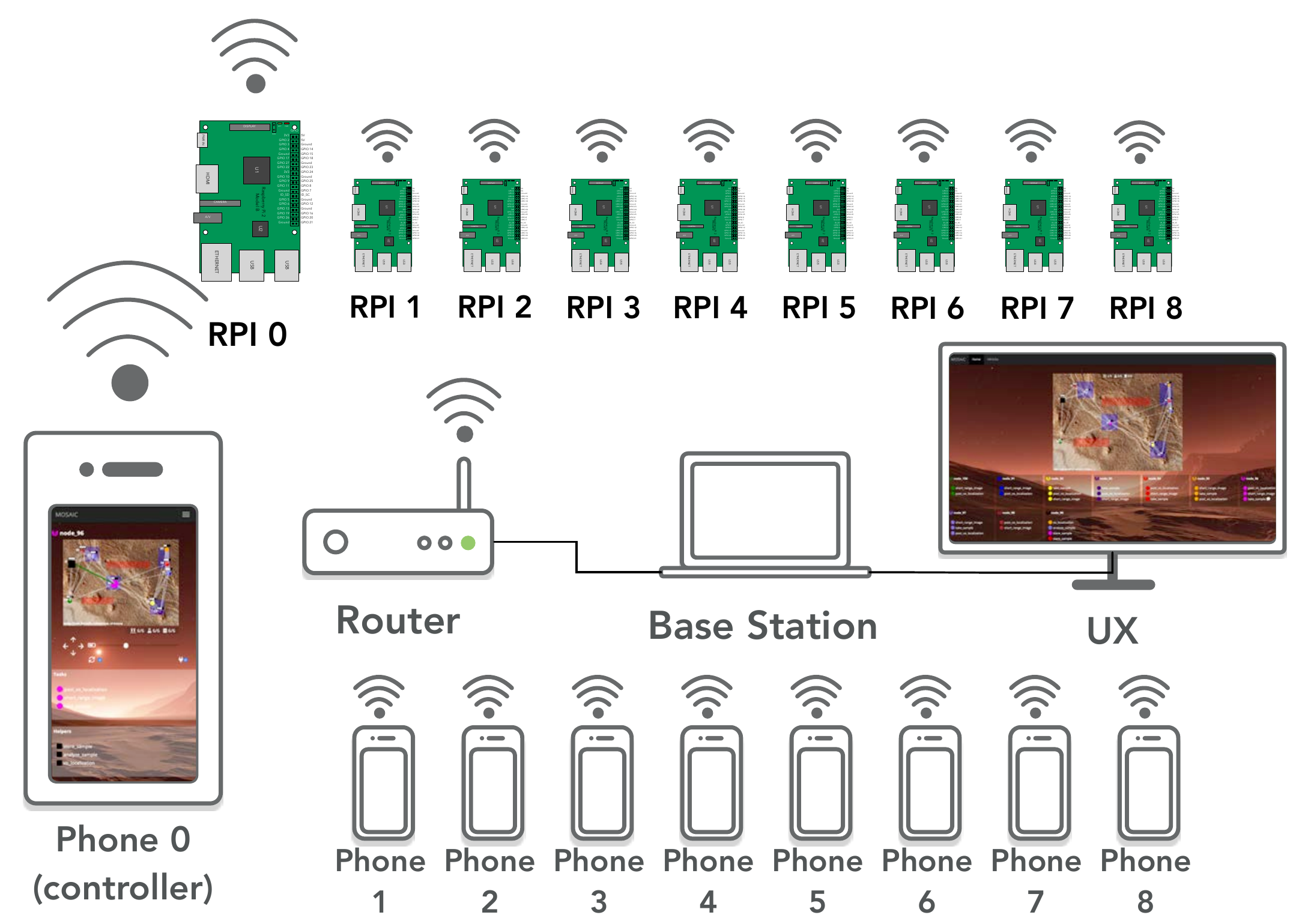}
\end{subfigure}
\caption{\emph{Top}: Live demonstration of PDRA at the 2019 ICAPS conference. \emph{Bottom}: system architecture. Attendees were invited to control the location and battery state of ten simulated robots through their portable devices. Each robot's PDRA and ROC was executed on a dedicated Raspberry Pi, while a laptop represented the base station. The ROCs reallocated tasks in real-time in response to the attendees' commands.}
\label{fig:experiments:icaps}
\end{figure}

In order to stress-test the performance of PDRA, we performed a human-in-the-loop demonstration of the software at the 2019 ICAPS conference (Figure~\ref{fig:experiments:icaps}). The PDRA software and task allocation algorithm were deployed on nine Raspberry Pi 3 embedded computers, each representing a robotic agent, aided by a laptop acting as a base station and communicating through WiFi. A web-based interface allowed conference attendees to change the simulated location of the robots, affecting network bandwidths and the robots' ability to collect samples, and their battery levels, affecting computational performance. The robots' task allocation algorithms re-allocated tasks in real-time in response to user inputs, and the operations of the system were displayed on a large screen.
The demonstration qualitatively demonstrated the robustness of overall architecture in presence of a variety of (occasionally mildly adversarial) system states and network topologies, and showcased its ability to perform well on highly computationally constrained architectures.

\jvspace{-.5em}

\section{Conclusions}
\label{sec:conclusions}

In this paper we presented the Pluggable Distributed Resource Allocator (PDRA), a pluggable middleware that enables networked mobile robots to share computational tasks across a time-varying network and ``plugs in''  to existing single-agent schedulers/executives and computation resources, allowing easy integration in existing autonomy architectures.

A number of directions for future work are of interest.
First, we plan to increase PDRA's robustness to disagreements in the robot's world views at the plan execution level, and to integrate monitoring tools to identify disagreements.
Second, we will study stochastic task allocation algorithms that are robust to disagreements in the world views and that can cope with unknown or partially-known contact plans.
\rev{Third, we will assess the performance of PDRA in conjunction with alternative task allocation algorithms such as distributed auctions and DCOP, and investigate the impact of these allocation mechanism on system-level performance for static and time-varying network topologies.}
Fourth, we plan to extend PDRA's applicability beyond pure functions to \emph{stateful} computational tasks (e.g., SLAM), where transferring a computational task between robots incurs a sizable one-off communication cost. 
Finally, we will develop user interfaces for human supervision and operation of PDRA, which are key to increasing operator trust and fostering the adoption of PDRA in real-world multi-robot exploration tasks. %

\section*{Acknowledgments}
The research was carried out at the Jet Propulsion Laboratory, California Institute of Technology, under a contract with the National Aeronautics and Space Administration.

\bibliographystyle{IEEEtran}
\bibliography{IEEEabrv,bibliography.bib}

\end{document}